\documentclass[journal]{IEEEtran}
\usepackage{cite}

\ifCLASSINFOpdf
  \usepackage[pdftex]{graphicx}
\else
  \usepackage[dvips]{graphicx}
\fi
\usepackage{amsmath}
\usepackage{amssymb} 
\usepackage{amsthm}

\DeclareMathOperator*{\argmax}{argmax}

\usepackage{algorithmic}

\usepackage{algorithm}

\usepackage{array}

\usepackage{threeparttable}
\usepackage{booktabs}

\ifCLASSOPTIONcompsoc
  \usepackage[caption=false,font=normalsize,labelfont=sf,textfont=sf]{subfig}
\else
  \usepackage[caption=false,font=footnotesize]{subfig}
\fi
\graphicspath{{updated_imgs/}}
\usepackage{pgf}
\usepackage{tikz}
\usetikzlibrary{arrows,automata}
\usetikzlibrary{shapes.geometric,shapes.arrows,decorations.pathmorphing}
\usetikzlibrary{matrix,chains,scopes,positioning,arrows,fit}

\begin{document}

\def\mi{\mathrm{MI}}
\def\vi{\mathrm{VI}}
\def\dist{\mathrm{TDAC}}
\def\divs{\mathrm{TDAS}}
\def\algpru{\emph{ALG}}
\def\distributed{\emph{EPFD}}
\def\greedy{\emph{COMEP}}
\def\ddismi{\emph{DOMEP}}
\def\mumble{\mathrm{mumble}}
\def\eg{e.g.,}
\def\ie{i.e.,}
\def\etal{\emph{et al.}}
\def\entitle{Ensemble Pruning based on Objection Maximization with a General Distributed Framework} 
\def\fulldist{a trade-off between diversity and accuracy of two individual classifiers}
\def\fulldivs{a trade-off between diversity and accuracy of a set of an ensemble}
\def\fullmypru{Objection Maximization for Ensemble Pruning} 
\def\fullgreedy{Centralized \fullmypru{}}
\def\fullddismi{Distributed \fullmypru{}}
\def\fullframe{Ensemble Pruning Framework in a Distributed Setting}

%
\title{\entitle{}}
%
%
%

\author{Yijun~Bian, 
        Yijun~Wang, 
        Yaqiang~Yao, 
        and~Huanhuan~Chen,~\IEEEmembership{Senior~Member,~IEEE}
\thanks{Y. Bian, Y. Wang, Y. Yao, and H. Chen were with the School of Computer Science and Technology, University of Science and Technology of China (USTC), Hefei 230027, China (e-mails: yjbian@mail.ustc.edu.cn; wyjun@mail.ustc.edu.cn; yaoyaq@mail.ustc.edu.cn; hchen@ustc.edu.cn). 
This work was supported in part by the National Key Research and Development Program of China under Grant 2016YFB1000905, and in part by the National Natural Science Foundation of China under Grants 91846111 and 91746209. Corresponding author: Huanhuan Chen. }
\thanks{Manuscript received January 07, 2019; revised June 23, 2019; accepted September 25, 2019.}}

%
%

\markboth{IEEE Transactions on Neural Networks and Learning Systems,~Vol.~, No.~, Month~Year}%
{Bian \MakeLowercase{\textit{et al.}}: \entitle{}}

%



\maketitle

\begin{abstract}
Ensemble pruning, selecting a subset of individual learners from an original ensemble, alleviates the deficiencies of ensemble learning on the cost of time and space. 
Accuracy and diversity serve as two crucial factors while they usually conflict with each other. 
To balance both of them, we formalize the ensemble pruning problem as an objection maximization problem based on information entropy. 
Then we propose an ensemble pruning method including a centralized version and a distributed version, in which the latter is to speed up the former. 
At last, we extract a general distributed framework for ensemble pruning, which can be widely suitable for most of the existing ensemble pruning methods and achieve less time consuming without much accuracy degradation. 
Experimental results validate the efficiency of our framework and methods, particularly concerning a remarkable improvement of the execution speed, accompanied by gratifying accuracy performance. 
\end{abstract}


%
\IEEEpeerreviewmaketitle

\section{Introduction}
\label{intro}

\IEEEPARstart{T}{hanks} to its remarkable potential, ensemble learning has attracted an amount of interest in the machine learning community \cite{zhou2012ensemble} and has been applied widely in many real-world tasks such as object detection, object recognition, and object tracking \cite{girshick2014rich,wang2012mining,zhou2014ensemble, ykhlef2017efficient}. 
As it is also known as committee-based learning, multiple classifier systems, or mixtures of experts \cite{zhou2012ensemble,kuncheva2003diversity,tang2006analysis}, an ensemble is a set of learned models that make decisions collectively rather than relying on one single model. 
The variety of types of individual learners categorizes an ensemble as heterogeneous ensembles and homogeneous ensembles. 
And most of the ensemble methods concentrate on the latter such as bagging \cite{breiman1996bagging} and boosting \cite{freund1995boosting,freund1996experiments}. 

The success of ensemble methods is commonly attributable to two key issues: the accuracy of individual learners and the diversity among them \cite{dietterich2000ensemble}. 
For classification problems, one classifier is accurate if its error rate is better than random guessing on new instances; two classifiers are diverse if they make different errors on new instances \cite{dietterich2000ensemble}. 
Unfortunately, there is still no consensus in the community on the definition or measurement for diversity, unlike the apparent accuracy. 
Besides, the diversity among individual classifiers usually decreases when these individuals approach a higher level of accuracy. 
Thus how to handle the trade-off between the two criteria is an essential issue in ensemble learning. 

Although ensemble methods are effectual, one significant drawback here is that both the required memory and processing time increase visibly with the number of individual learners in the ensemble. 
To mitigate this shortcoming motivates ensemble pruning that aims to select a subset of individual learners in an ensemble, named as ensemble selection or ensemble thinning as well \cite{liu2014ensemble,kokkinos2015confidence,zhang2017ranking,lysiak2014optimal,partalas2008focused,caruana2004ensemble, banfield2005ensemble,chen2009predictive}. 
It could even improve the generalization performance of an ensemble with a smaller size \cite{zhou2002ensembling}. 
There has been a numerous progression on ensemble pruning methods in the last two decades. 
Most of the existing pruning methods, however, are centralized in which all individual classifiers have to be stored and processed on one single machine. 
As the scale of data and an ensemble itself enlarges rapidly in the context of big data, the performance of centralized methods is becoming the bottleneck in execution time, which is why distributed approaches need to emerge. 

To deal with ensemble pruning problems fast with balancing diversity and accuracy appropriately, we firstly treat ensemble pruning as an objection maximization problem using information entropy to reflect diversity and accuracy. 
The objective function that we aim to maximize is a trade-off between diversity and accuracy from an information entropy perspective. 
Secondly, we transform this approach to one distributed version to speed up the execution, inspired by the emerging concept of ``composable core-sets'' in recent years. 
It adopts the same idea as a two-round divide-and-conquer strategy, which is particularly suitable for distributed settings. 
Thirdly, we extract a general distributed framework for ensemble pruning from our method's distributed version. 
It could be widely applicable to various existing methods for ensemble pruning and achieve less time consuming without much accuracy degradation. 

Our contribution in this paper is four-fold: 
\begin{itemize}
    \item We formalize the ensemble pruning problem as an objection maximization problem based on information entropy, in order to balance diversity and accuracy. 
	\item We propose an ensemble pruning method including a centralized version and a distributed version, utilizing accuracy and diversity concurrently.
	\item We propose a general distributed framework for ensemble pruning, which could be widely utilized and achieve less time consuming without much accuracy degradation. 
	\item We design detailed experiments to validate the effectiveness of our distributed framework and approaches.
\end{itemize}

\section{Related Work}
\label{related}

In this section, we introduce diversity (\ie{} a key concept in ensemble learning) and existing research on it firstly. 
Then we describe the difficulty and existing methods in ensemble pruning. 
Thirdly, we introduce a concept of ``composable core-sets'' and its development which sheds some light on our work. 
Finally, we explain the distinction between our proposed methods and other existing ensemble pruning method and summarize our contribution specifically. 

\subsection{Diversity in Ensemble Learning}

Diversity, intuitively considered as the difference among individual learners in an ensemble, is a fundamental issue in ensemble methods \cite{zhou2012ensemble}, with several alternative names as dependence, orthogonality or complementarity of learners \cite{kuncheva2003diversity}. 
Practically, individual classifiers are usually trained on the subsets of the same training data, which drives them highly correlated, breaks the assumption about the independence of individual classifiers, and makes it hard to seek diversity. 
Numerous ensemble methods attempt to encourage diversity implicitly or heuristically \cite{yu2011diversity}. 
For instance, boosting and bagging promote diversity by re-weighting and sub-sampling existing training samples, respectively \cite{breiman1996bagging, freund1995boosting,freund1996experiments, melville2003constructing, soares2017cluster}; neural networks (NN) ensembles also create diversity using different initial weights, different architectures of the networks, and different learning algorithms. 

Unfortunately, researchers still have not reached a consensus yet on an official measure of diversity. 
Several measures have been proposed to represent diversity, which could be generally divided into pairwise and non-pairwise diversity measures~\cite{kuncheva2003diversity}, while no superior exists \cite{tang2006analysis}. 
Based on coincident errors between a pair of individual classifiers, pairwise diversity represents the behavior whether both of them predict an instance identically or disagree with each other, including $Q$-statistic~\cite{yule1900association}, $\kappa$-statistic \cite{cohen1960coefficient}, disagreement measure \cite{skalak1996sources,ho1998random}, correlation coefficient \cite{sneath1973numerical}, and double-fault \cite{giacinto2001design}. 
In contrast, non-pairwise diversity is the average of all possible pairs, directly measuring a set of classifiers using the variance, entropy, or the proportion of individual classifiers that fail on instances chosen randomly. 
It includes interrater agreement~\cite{fleiss2013statistical}, Kohavi-Wolpert variance \cite{kohavi1996bias}, the entropy of the votes \cite{cunningham2000diversity,shipp2002relationships}, the difficulty index \cite{kuncheva2003diversity,hansen1990neural}, the generalized diversity~\cite{partridge1997software}, and the coincident failure diversity~\cite{partridge1997software}. 
Besides, there are also two other measures that do not fall into these two mentioned categories previously: 
One is the correlation penalty function~\cite{liu1999ensemble}, measuring the diversity of each member against the entire ensemble in the negative correlation learning (NCL) \cite{chen2009regularized,chen2010multiobjective}; The other is ambiguity~\cite{zenobi2001using}, measuring the average offset of each member against the entire ensemble output. 

Moreover, few researchers could tell how diversity works exactly although the crucial role of diversity has been widely accepted in ensemble methods. 
In the last decade or so, Brown~\cite{brown2009information} claimed that from an information theoretic perspective, diversity within an ensemble existed indeed on numerous levels of interaction between the classifiers. 
His work inspired Zhou and Li~\cite{zhou2010multi} to propose that the mutual information should be maximized to minimize the prediction error of an ensemble from the view of multi-information. 
Subsequently, Yu \etal{}~\cite{yu2011diversity} claimed that the diversity among individual learners in a pairwise manner, used in their diversity regularized machine (DRM), could reduce the hypothesis space complexity, which implied that controlling diversity played the role of regularization in ensemble methods.

\subsection{Ensemble Pruning}

Ensemble pruning deals with the reduction of an ensemble while improving its efficiency and predictive performance~\cite{tsoumakas2009ensemble}. 
Margineantu and Dietterich \cite{margineantu1997pruning} showed the possibility to obtain nearly the same level of performance as the entire set by selecting a subset of learners from an ensemble in the first study on ensemble pruning. 
Zhou \etal{}~\cite{zhou2002ensembling} provided the bias-variance decomposition of error as the principal factor of the success of their approach named as ``Genetic Algorithm based on Selective Ensemble (GASEN)'', and claimed that pruning could lead to smaller ensembles with better generalization performance. 
It is difficult, however, to select the sub-ensembles with the best generalization performance. 
One trouble is to estimate the generalization performance of a sub-ensemble, and the other is that finding the optimal subset is a combinatorial search problem with exponential computational complexity \cite{li2012diversity}. 
Note that selecting the best combination of classifiers from an ensemble is NP-complete hard and even intractable to approximate \cite{martinez2007using}. 

Numerous ensemble pruning methods have been proposed to overcome shortcomings of ensemble learning over the last two decades, which could be categorized into three general families: ranking-based, clustering-based, and optimization-based. 
Ranking-based pruning methods, the simplest conceptually, order the learners in the ensemble and select the first few of them according to different evaluation functions \cite{tsoumakas2009ensemble}, including minimizing the error (\eg{} Orientation Ordering~\cite{martinez2006pruning}), maximizing the diversity (\eg{} KL-divergence Pruning and Kappa Pruning \cite{margineantu1997pruning}), or combining them both (\eg{} Diversity Regularized Ensemble Pruning \cite{li2012diversity}). 
Clustering-based pruning methods employ a clustering algorithm to detect groups of learners that make similar predictions initially and then prune each cluster separately to increase the overall diversity of the ensemble \cite{tsoumakas2009ensemble}. 
Note that an intrinsic property that those methods could be executed in a parallel manner is ignored frequently in the second phase. 
Optimization-based pruning methods pose ensemble pruning as an optimization problem which is to find the subset of the original ensemble that optimizes a measure indicating its generalization performance. 
Searching exhaustively in the space of ensemble subsets is unfeasible even for a moderate ensemble size since this problem is NP-complete hard \cite{li2012diversity,martinez2007using}. 
Thus various techniques are utilized to alleviate this predicament including genetic algorithm \cite{zhou2003selective}, greedy algorithm \cite{partalas2012study}, hill climbing \cite{guo2016novel}, and bi-objective evolutionary optimization \cite{qian2015pareto}.

\subsection{Composable Core-sets}

Over the last few years, an effective technique, captured via the concept of ``composable core-sets'', arises in order to solve optimization problems over large data sets in the distributed computing literature. 
Its effectiveness has been confirmed empirically for many machine learning applications, such as diverse nearest neighbor search \cite{indyk2014composable}, diversity maximization~\cite{aghamolaei2015diversity}, and feature selection \cite{zadeh2017scalable}. 

The notion of ``composable core-sets'' is introduced explicitly by Indyk \etal{}~\cite{indyk2014composable} for the very first time, while the notion of ``core-sets'' can be dated back to \cite{agarwal2004approximating}. 
A core-set for an optimization problem, informally, is a subset (with a guaranteed approximation factor) of that data on which solving the underlying problem could yield an approximate solution for the original data. 
Composable core-sets are a collection of core-sets in which the union of them gives a core-set for the union of the original data subsets \cite{indyk2014composable}. 
Besides, a composable core-set with $\alpha$ approximate factor yields a solution which is an approximation of the optimal solution for the optimization problem, and the approximation is guaranteed by a factor $\alpha$, which is $1/12$ \cite{aghamolaei2015diversity} and could be improved to $8/25$ \cite{zadeh2017scalable}.

\subsection{Our Contribution}

An essential distinction between our proposed methods and other existing ensemble pruning methods is that our methods could tackle the ensemble pruning task in a distributed way, accelerating the pruning process substantially. 
Instead of using any previous measure of diversity, our methods utilize the concept of the mutual information $\mathrm{I}(\cdot;\cdot)$ \cite{cover2012elements}, the normalized mutual information $\mi(\cdot,\cdot)$, and the normalized variation information $\vi(\cdot,\cdot)$ \cite{zadeh2017scalable} from an information entropy perspective, to conduct an objective function and take diversity and accuracy into consideration implicitly and simultaneously. 
Besides, our proposed framework (\distributed{}) could be widely applied to various existing ensemble pruning methods, to achieve less time consuming without much accuracy degradation, which is an impressive advantage of our method.

\section{Methodology}
\label{methods}

In this section, we firstly elaborate our objection maximization based on information entropy for ensemble pruning in a centralized way, then attain a distributed version by introducing the concept of composable core-sets, and finally extract a general distributed framework for ensemble pruning.

\subsection{Objection Maximization Based on Information Entropy for Ensemble Pruning}

Given a large data set $\mathcal{D}$ with the size $d$ of labeled instances obtained gradually from stream data, and their labels represented by a $d$-dimensional vector $\mathbf{c}$. 
A set of $n$ trained individual classifiers $\mathcal{H}=\{h_i\}_{i=1}^n$ is considered as the original ensemble, in which each one maps the feature space of instances to the label space. 
The classification result vector of any individual classifier $h_i$ from the ensemble $\mathcal{H}$ on the data set $\mathcal{D}$, similar to the class label vector $\mathbf{c}$, is represented by a $d$-dimensional vector $\mathbf{h}_i$. 
The ensemble pruning task aims to find a compact subset of the original ensemble which will predict the labels with high accuracy. 
These chosen individual classifiers need to be diverse and accurate simultaneously to achieve this goal. 
To this end, we select some diversified individual classifiers from the original ensemble which are relevant to the vector of class labels. 
Hence we strive to define a metric distance between individual classifiers in consideration of diversity and accuracy concurrently inspired by \cite{zadeh2017scalable} so that the ensemble pruning problem would be reduced to an objection maximization problem.

Given two discrete random variables $X$ and $Y$, Cover and Thomas~\cite{cover2012elements} defined the mutual information $\mathrm{I}(\cdot;\cdot)$ between them, \ie{} 
\begin{equation}
\small
\begin{split}
    \mathrm{I}(X;Y) &= \mathrm{H}(X) - \mathrm{H}(X|Y) \\
    &= \sum_{x\in X, y\in Y} p(x,y)\log\frac{p(x,y)}{p(x)p(y)} \,,
\end{split}
\end{equation}
and then Zadeh \etal{}~\cite{zadeh2017scalable} defined the normalized mutual information $\mi{}(\cdot,\cdot)$ and the normalized variation of information $\vi{}(\cdot,\cdot)$ of them, \ie{} 
\begin{equation}
\small
\begin{split}
    \mi{}(X,Y) = 
    \frac{\mathrm{I}(X;Y)}{\sqrt{\mathrm{H}(X)\mathrm{H}(Y)}} \,,
\end{split}
\end{equation}
\begin{equation}
\small
\begin{split}
    \vi{}(X,Y) = 1- 
    \frac{\mathrm{I}(X;Y)}{\mathrm{H}(X,Y)} \,,
\end{split}
\end{equation}
wherein $p(\cdot,\cdot)$, $\mathrm{H}(\cdot)$, and $\mathrm{H}(\cdot,\cdot)$ are the joint probability, the entropy function, and the joint entropy function, respectively.

Consider the class label vector $\mathbf{c}$ and two classification result vectors ($\mathbf{h}_i$ and $\mathbf{h}_j$) generated over the data set $\mathcal{D}$ by any two individual classifiers ($h_i$ and $h_j$). 
In this case, 
the normalized mutual information $\mi{}(\mathbf{h}_i,\mathbf{c})$ exhibits the relevance between this individual classifier $h_i$ and the class label vector $\mathbf{c}$, implying the accuracy of this individual classifier on the training data set; 
the normalized variation of information $\vi{}(\mathbf{h}_i,\mathbf{h}_j)$ reveals the redundancy between these two individual classifiers, indicating the diversity between them. 
Since class labels have already been discrete values and these values are only relevant to the number of classes in those used data sets, we do not need to discretize continuous variables to calculate the probabilities used in $\mi{}(\cdot,\cdot)$ and $\vi{}(\cdot,\cdot)$, while Zadeh \etal{}~\cite{zadeh2017scalable} have to deal with it.

In order to take both diversity and accuracy into consideration concurrently, the objective function between two individual classifiers (\emph{\fulldist}, $\dist$) is defined naturally as 
\begin{equation}
\small
\begin{split}
    & \dist{}(h_i,h_j) \\= & 
    \begin{cases}
        \lambda \vi(\mathbf{h}_i,\mathbf{h}_j) + (1-\lambda)\frac{ \mi(\mathbf{h}_i,\mathbf{c}) + \mi(\mathbf{h}_j,\mathbf{c}) }{2} , & \text{if } h_i\neq h_j \,;\\
        0 , & \text{otherwise} \,, 
    \end{cases}
\end{split}
\label{eq:4}
\end{equation}
where a regularization factor $\lambda$ is introduced to balance between these two criteria, indicating their importance as well. 
The first criterion is to raise diversity by avoiding redundancy, and the second one is to promote accuracy by maximizing their relevance. 
Note that $\vi{}(\cdot,\cdot)$ is metric \cite{vinh2010information} and $\mi{}(\cdot,\mathbf{c})$ is non-negative \cite{zadeh2017scalable}. 
Consequently $\dist{}(\cdot,\cdot)$ is metric as well, which means 
$\dist(h_i,h_j)+\dist(h_j,h_k) \geqslant\dist(h_i,h_k)$.

Subsequently, for an ensemble $\mathcal{H}$ that is a set composed of $n$ individual classifiers, the objection (\emph{\fulldivs}, $\divs$) is defined naturally as 
\begin{equation}
\small
    \divs(\mathcal{H}) = \frac{1}{2} \sum_{h_i\in\mathcal{H}} \sum_{h_j\in\mathcal{H}} \dist(h_i,h_j) 
    \,. \label{divs1}
\end{equation}
Note that $\divs(\mathcal{H})$ in Eq.~(\ref{divs1}) could be reformulated as 
\begin{equation}
\small
\begin{split}
    & \divs(\mathcal{H}) \\= &
    \frac{1}{2}\lambda \sum_{h_i\in\mathcal{H}} \sum_{h_j\in\mathcal{H}} \vi(\mathbf{h}_i,\mathbf{h}_j) + 
    \frac{n-1}{2}(1-\lambda) \sum_{h_i\in\mathcal{H}} \mi(\mathbf{h}_i,\mathbf{c}) \,,
\end{split}
\label{divs2}
\end{equation}
where $\vi(\cdot,\cdot)$ of two similar individual classifiers will be near to zero. 
The first term in Eq.~(\ref{divs2}) prevents to select similar individual classifiers, and the second term ensures that those selected individual classifiers are relevant to the class labels. 
Therefore the ensemble pruning task is reformulated as an objective function maximization problem, which aims to find a subset $\mathcal{P}\subset\mathcal{H}$ with a specified condition $|\mathcal{P}|=k$ to restrict the size of the pruned sub-ensemble, 
\begin{equation}
\small
    \max_{\substack{\scriptscriptstyle \mathcal{P}\subset\mathcal{H} ,\\ |\mathcal{P}|=k}} \divs(\mathcal{P}) = 
    \max_{\substack{\scriptscriptstyle \mathcal{P}\subset\mathcal{H} ,\\ |\mathcal{P}|=k}} \frac{1}{2} \sum_{h_i\in\mathcal{P}} \sum_{h_j\in\mathcal{P}} \dist(h_i,h_j) 
    \,. \label{divs}
\end{equation}

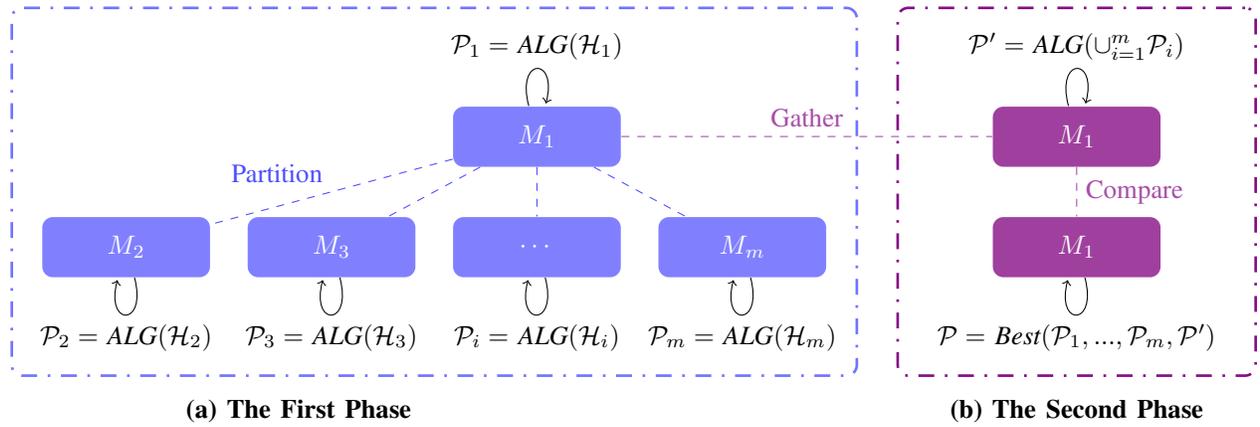
\begin{figure*}[t]
\centering
\begin{minipage}{0.926\linewidth}
\begin{tikzpicture}
    \matrix (m) [matrix of nodes, 
        column sep=5mm, 
        row sep=6.7mm,
        nodes={
            fill=blue!50, text=white, 
            anchor=center, 
            text centered, 
            rounded corners, 
            minimum width=2.23cm, minimum height=8mm
        }, 
        right iso/.style={isosceles triangle,scale=0.5,sharp corners, anchor=center, xshift=-4mm}, 
        left iso/.style={right iso, rotate=180, xshift=-8mm}, 
        txt/.style={text width=1.5cm,anchor=center}, 
        ellip/.style={ellipse,scale=0.5}, 
        empty/.style={draw=none}
    ]
    {
            |[fill=white,minimum height=0.5mm]| $~$   
        &   |[fill=white,minimum height=0.5mm]| $~$   
        &   |[fill=white,minimum height=0.5mm]| $~$   
        &   |[fill=white,minimum height=0.5mm]| $~$   
        &   |[fill=white,minimum width=0.5mm,minimum height=0.5mm]| $~$  
        &   |[fill=white,minimum width=0.5mm,minimum height=1.5mm]| $~$
        &   |[fill=white,minimum height=0.5mm]| $~$   
        &   |[fill=white,minimum width=0.5mm,minimum height=1.5mm]| $~$  
        \\
            |[fill=white]| $~$   
        &   |[fill=white]| $~$   
        &   $M_1$    
        &   |[fill=white]| $~$   
        &   |[fill=white,minimum width=0.5mm,minimum height=0.5mm]| $~$  
        &   |[fill=white,minimum width=0.5mm,minimum height=1.5mm]| $~$  
        &   |[fill=violet!75]| $M_1$ 
        &   |[fill=white,minimum width=0.5mm,minimum height=1.5mm]| $~$  
        \\
            $M_2$ & $M_3$ & $\cdots$ & $M_m$ 
        &   |[fill=white,minimum width=0.5mm,minimum height=0.5mm]| $~$  
        &   |[fill=white,minimum width=0.5mm,minimum height=1.5mm]| $~$  
        &   |[fill=violet!75]| $M_1$ 
        &   |[fill=white,minimum width=0.5mm,minimum height=1.5mm]| $~$  
        \\
            |[fill=white,minimum height=0.5mm]| $~$   
        &   |[fill=white,minimum height=0.5mm]| $~$   
        &   |[fill=white,minimum height=0.5mm]| $~$   
        &   |[fill=white,minimum height=0.5mm]| $~$   
        &   |[fill=white,minimum width=0.5mm,minimum height=0.5mm]| $~$  
        &   |[fill=white,minimum width=0.5mm,minimum height=0.5mm]| $~$  
        &   |[fill=white,minimum height=0.5mm]| $~$   
        &   |[fill=white,minimum width=0.5mm,minimum height=0.5mm]| $~$  
        \\
    };
    
    \path   (m-2-3) edge [loop above] node {$\mathcal{P}_1=$ \algpru{}$(\mathcal{H}_1)$} (m-2-1)
            (m-3-1) edge [loop below] node {$\mathcal{P}_2=$ \algpru{}$(\mathcal{H}_2)$} (m-3-1)
            (m-3-2) edge [loop below] node {$\mathcal{P}_3=$ \algpru{}$(\mathcal{H}_3)$} (m-3-2)
            (m-3-3) edge [loop below] node {$\mathcal{P}_i=$ \algpru{}$(\mathcal{H}_i)$} (m-3-3)
            (m-3-4) edge [loop below] node {$\mathcal{P}_m=$ \algpru{}$(\mathcal{H}_m)$} (m-3-4);
    \path   (m-2-7) edge [loop above] node {$\mathcal{P}'=$ \algpru{}$(\cup_{i=1}^m\mathcal{P}_i)$} (m-2-7)
            (m-3-7) edge [loop below] node {$\mathcal{P}=$ \emph{Best}$(\mathcal{P}_1,...,\mathcal{P}_m, \mathcal{P}')$} (m-3-7);
    \path[-, dashed] (m-2-3) edge [above left,color=blue!70] node {Partition} (m-3-1)
                     edge [above left,color=blue!70] node {} (m-3-2)
                     edge [above left,color=blue!70] node {} (m-3-3)
                     edge [above left,color=blue!70] node {} (m-3-4);
    \path[-, dashed] (m-2-3) edge [above,color=violet!70] node {Gather } (m-2-7)
             (m-2-7) edge [right,color=violet!70] node {Compare} (m-3-7);
    
\tikzset{blue dotted/.style={draw=blue!55!white, line width=1pt,
                             dash pattern=on 1pt off 4pt on 6pt off 4pt,
                             inner sep=4mm, rectangle, rounded corners}};
\tikzset{violet dotted/.style={draw=violet!95!white, line width=1pt,
                               dash pattern=on 1pt off 4pt on 6pt off 4pt,
                               inner sep=4mm, rectangle, rounded corners}};

\node (first dotted box) [blue dotted, 
                          fit = (m-1-1) (m-4-4)] {};
\node (second dotted box) [violet dotted,
                          fit = (m-1-6) (m-4-8)] {};

\node at (first dotted box.south) [below left, inner sep=3mm] {\textbf{(a) The First Phase}};
\node at (second dotted box.south) [below, inner sep=3mm] {\textbf{(b) The Second Phase}};
\end{tikzpicture}
\end{minipage}
\caption{ 
Diagram of Algorithm~\ref{distributed}: \fullframe{} (\distributed{}). 
(a) In the first phase, a primary machine (e.g., $M_1$) partitions all individual classifiers in the original ensemble into $m$ groups $\{\mathcal{H}_i\}_{i=1}^m$ randomly and allocates them to different machines. For each $i\,(1\leqslant i\leqslant m)$, the machine $M_i$ runs \algpru{} on its allocated set $\mathcal{H}_i$ independently and selected a subset $\mathcal{P}_i$ from it in parallel. 
(b) In the second phase, the primary machine (e.g., $M_1$) gathers all subsets, runs \algpru{} on their union $\cup_{i=1}^m\mathcal{P}_i$ to produce a subset $\mathcal{P}'$, and eventually outputs the best one of them by comparing $\mathcal{P}'$ with $\mathcal{P}_i \, (1\leqslant i\leqslant m)$.  
Remark: (1) \algpru{} could be any one of existing pruning method, including \greedy{}. (2) The final $\mathcal{P}$ is chosen by $\textit{Best}(\cdot)$ according to some certain criteria such as accuracy or $\divs{}(\cdot)$ in \ddismi{}.  
}\label{fig:framework}
\end{figure*}

It is a bi-objective optimization problem to combine diversity and accuracy concurrently for ensemble pruning tasks, but we could manage to reformulate it as a single-objective optimization problem by introducing $\dist{}(\cdot,\cdot)$ in Eq.~(\ref{eq:4}). 
That is one way to solve the problem by using the objective weighting \cite{srinivas1994muiltiobjective}. 
Another way could be introducing a concept of domination~\cite{deb2002fast,qian2015pareto} to obtain a Pareto optimal solution, which we will leave it for future work. 
Besides, our solution manages to fix the size of the pruned sub-ensemble while the latter solution cannot, which usually leads to oversize sub-ensembles and affects the space cost.

\begin{algorithm}[h]
\small
\caption{\small \fullgreedy{} (\greedy{})}
\label{greedy}
\begin{algorithmic}[1]
    \REQUIRE Set of an original ensemble $\mathcal{H}$, threshold $k$ as the size of the pruned sub-ensemble. 
    \ENSURE Set of the pruned sub-ensemble $\mathcal{P}$ satisfying that $\mathcal{P}\subset\mathcal{H}$ and $|\mathcal{P}|\leqslant k$. 
    \STATE $\mathcal{P}\gets$ an arbitrary individual classifier $h_i\in\mathcal{H}$. 
    \FOR{$2\leqslant i\leqslant k$}
        \STATE $h^*\gets \argmax_{h_i\in\mathcal{H}\setminus\mathcal{P}} \sum_{h_j\in\mathcal{P}} \dist(h_i,h_j)$. 
        \STATE Move $h^*$ from $\mathcal{H}$ to $\mathcal{P}$. 
    \ENDFOR
\end{algorithmic}
\end{algorithm}

Up to now, we have modeled the ensemble pruning task through an objective function maximization problem as shown in Eq.~(\ref{divs}), which is enough to form a centralized algorithm to accomplish this goal of ensemble pruning. 
This centralized method, named as ``\emph{\fullgreedy{} (}\greedy{}\emph{)}'' shown in Algorithm~\ref{greedy}, selects greedily the current optimal classifier at each step, and could achieve a $1/2$ approximation factor for objective function maximization problem according to \cite{birnbaum2009improved}.

\subsection{\fullddismi}

``\emph{\fullddismi{} (}\ddismi{}\emph{)}'', the distributed version of \greedy{} shown in Algorithm~\ref{ddismi}, adopts a two-round divide-and-conquer strategy and composable core-sets~\cite{mirrokni2015randomized} as guidelines, which are particularly suitable for distributed settings. 
It partitions a set of individual classifiers of an ensemble into smaller pieces, solves the ensemble pruning problem on each piece separately, and eventually obtains a subset from the union of these representative subsets for all pieces.

Consider a set of $n$ trained individual classifiers $\mathcal{H}=\{h_i\}_{i=1}^n$ as the original ensemble. 
In the first phase, a primary machine partitions all individual classifiers in the original ensemble into $m$ groups $\{\mathcal{H}_i\}_{i=1}^m$ randomly and allocates them to different machines. 
Note that $\cup_{i=1}^m\mathcal{H}_i=\mathcal{H}$ where $m$ is the number of machines, and that the primary machine could be any one of these available machines. 
For each $i~(1\leqslant i\leqslant m)$, machine $i$ runs \greedy{} on its allocated set $\mathcal{H}_i$ independently and selects a subset $\mathcal{P}_i$ from it in parallel. 
In the second phase, the primary machine gathers all subsets, runs \greedy{} on their union $\cup_{i=1}^m \mathcal{P}_i$ to produce a subset $\mathcal{P}'$, and eventually outputs the best one of them by comparing $\mathcal{P}'$ with $\mathcal{P}_i ~(1\leqslant i\leqslant m)$ according to Eq.~(\ref{divs}). 
It suffices to output the satisfying subset $\mathcal{P}$ after these two phases (Lines \ref{alg2:s1}--\ref{alg2:s5} in Algorithm~\ref{ddismi}) in practice and the additional comparison purposes to get a higher approximation factor which is $1/4$ theoretically and could even reach $8/25$ under some special conditions \cite{zadeh2017scalable}.

\begin{algorithm}[h]
\small
\caption{\small \fullddismi{} (\ddismi{})}
\label{ddismi}
\begin{algorithmic}[1]
    \REQUIRE Set of an original ensemble $\mathcal{H}$, threshold $k$ as the size of the pruned sub-ensemble, number of machines $m$. 
    \ENSURE Set of the pruned sub-ensemble $\mathcal{P}$ meeting that $\mathcal{P}\subset\mathcal{H}$ and $|\mathcal{P}|\leqslant k$. 
    \STATE Partition $\mathcal{H}$ randomly into $m$ groups as equally as possible, \ie{} 
    $\mathcal{H}_1, ..., \mathcal{H}_m$. 
    \label{alg2:s1}
    \FOR{$1\leqslant i\leqslant m$}
        \STATE $\mathcal{P}_i\gets$ \greedy{}($\mathcal{H}_i$, $k$). 
    \ENDFOR
    \STATE $\mathcal{P}'\gets$ \greedy{}($\cup_{i=1}^m\mathcal{P}_i$, $k$). 
    \label{alg2:s5}
    \STATE $\mathcal{P}\gets \argmax_{\mathcal{T}\in\{ \mathcal{P}_1,...,\mathcal{P}_m,\mathcal{P}' \}} \divs(\mathcal{T})$. 
    \label{alg2:s6}
\end{algorithmic}
\end{algorithm}

At last, there might be some confusion over how to partition $n$ individual classifiers in $\mathcal{H}$ randomly into $m$ groups as equally as possible. 
The first step is to shuffle these individual classifiers with a different random order, and the second step is to arrange them to get $m$ groups. 
There would be no doubt when $n$ is a multiple of $m$ in which each group would contain $n/m$ individual classifiers. 
But if $n$ is not a multiple of $m$, there would be $(n \mod m)$ groups of them that each contains $\lceil n/m \rceil$ individual classifiers and $(n\, \mumble{} \,m)$ groups of them that each contains $\lfloor n/m \rfloor$ individual classifiers. 
Notice that $n\, \mumble{} \,m = m\lceil n/m \rceil -n$ according to \cite{graham2012concrete}.

\subsection{A General Distributed Framework for Ensemble Pruning}

A general distributed framework is extracted from \ddismi{}, named as ``\emph{\fullframe{} (}\distributed{}\emph{)}'' shown in Algorithm~\ref{distributed}, which likewise adopts the two-round divide-and-conquer strategy and composable core-sets \cite{mirrokni2015randomized}. 
It enables the ensemble pruning problem to be solved fast in a distributed way. 
\begin{algorithm}[h]
\small
\caption{\small \fullframe{} (\distributed{})}
\label{distributed}
\begin{algorithmic}[1]
    \REQUIRE Set of an original ensemble $\mathcal{H}$, number of machines $m$, a pruning method \algpru{}. 
    \ENSURE Set of the pruned sub-ensemble $\mathcal{P}$ meeting that $\mathcal{P}\subset\mathcal{H}$. 
    \STATE Partition $\mathcal{H}$ into $\{\mathcal{H}_i\}_{i=1}^m$ randomly. 
    \FOR{$1\leqslant i\leqslant m$}
        \STATE $\mathcal{P}_i\gets$ output from any pruning method \algpru{} on $\mathcal{H}_i$. 
        \label{alg3:s3}
    \ENDFOR
    \STATE $\mathcal{P}'\gets$ output from \algpru{} on $\cup_{i=1}^m \mathcal{P}_i$. 
    \label{alg3:s5}
    \STATE $\mathcal{P}\gets$ the best one among $\mathcal{P}_i,...,\mathcal{P}_m,$ and $\mathcal{P}'$ according to some certain criteria such as accuracy. 
    \label{alg3:s6} 
\end{algorithmic}
\end{algorithm}
Ensemble pruning is usually described as a process to acquire the optimum subset from the original ensemble. 
Let \algpru{} denote an arbitrary algorithm to perform this task and $\mathcal{H}$ the original ensemble. 
\distributed{} consists of two main phases just like \ddismi{} that could be regarded as a special case of \distributed{} where \greedy{} is chosen as the used pruning method (Lines \ref{alg3:s3},\ref{alg3:s5} in Algorithm \ref{distributed}). 
Another key difference is that the criterion (Line \ref{alg3:s6} in Algorithm \ref{distributed}) here is not limited to Eq.~(\ref{divs}). 
For instance, it could use accuracy or any other measures corresponding to data to compare different subsets. 
\distributed{} is a simple yet powerful tool to accelerate the original methods for ensemble pruning without much performance degradation, which is elaborated in Section~\ref{cmp:distributed}.

\section{Experiments}
\label{experts}

In order to evaluate our proposed methods, in this section, we elaborate our experiments on 17 binary and 12 multi-class data sets including an image data set with 12,500 pictures (Dogs vs. Cats\footnote{http://www.kaggle.com/c/dogs-vs-cats}) and 28 data sets from UCI repository \cite{Lichman:2013}. 
Standard 5-fold cross-validation is used in these experiments where the entire data set is split into three parts in each iteration, with 60\% as the training set, 20\% as the validation set, and 20\% as the test set. 
Besides, we construct homogeneous ensembles using Bagging on various types of classifiers including decision trees (DT), naive Bayesian (NB) classifiers, $k$-nearest neighbors (KNN) classifiers, linear model (LM) classifiers, and linear SVMs (LSVM). 
An ensemble is firstly trained on the training set, then pruned by a pruning method on the validation set, and finally tested on the test set. 
The baselines that we consider are a variety of ranking-based methods as well as optimization-based methods. 
These ranking-based methods include KL-divergence Pruning (KL), Kappa Pruning (KP) \cite{margineantu1997pruning}, Orientation Ordering Pruning (OO) \cite{martinez2006pruning}, Reduce-Error Pruning (RE) \cite{martinez2009analysis}, Diversity Regularized Ensemble Pruning (DREP) \cite{li2012diversity}, and Ordering-based Ensemble Pruning (OEP); 
These optimization-based methods are Single-objective Ensemble Pruning (SEP) and Pareto Ensemble Pruning (PEP) \cite{qian2015pareto}. 
Note that several methods cannot fix the number of learners after ensemble pruning (such as OO, DREP, SEP, OEP, and PEP), while others could fix it by giving a pruning rate which is the up limit of the percentage of those discarded individual classifiers in the original ensemble. 
Those methods that cannot fix the size might lead to oversize or undersize sub-ensembles and affect their space cost. 
Due to space constraints, we only report the comparisons of the time cost and the test accuracy hereinafter.

\begin{table*}[t]
\centering
\caption{
Comparison of the state-of-the-art methods with \greedy{} and \ddismi{} using Bagging to produce an ensemble with DTs as individual classifiers.
}\label{tab:Bagging:DT}
\scalebox{0.97}{
\setlength{\tabcolsep}{1.6pt}
\begin{threeparttable}[b]
\begin{tabular}{c cccccccc cc}
    \toprule
    Dataset & KL & KP & OO & RE & DREP & SEP & OEP & PEP & \greedy{} & \ddismi{} \\
    \midrule
Iono	        &	89.43$\pm$3.44	&	90.29$\pm$3.41	&	90.57$\pm$4.69	&	88.57$\pm$3.03	&	89.71$\pm$3.70	&	90.86$\pm$2.96	&	89.71$\pm$4.33	&	\textbf{91.71$\pm$3.41}	&	91.14$\pm$3.41	&	91.71$\pm$3.56	\\
Liver	        &	61.45$\pm$8.24	&	62.32$\pm$5.12	&	64.06$\pm$7.35	&	61.16$\pm$4.96$\ddagger$	&	56.23$\pm$8.41	&	60.87$\pm$4.35$\ddagger$	&	64.64$\pm$4.54	&	62.32$\pm$4.23	&	62.90$\pm$4.18	&	\textbf{64.64$\pm$4.30}	\\
Spam	        &	93.28$\pm$1.13	&	93.21$\pm$0.93	&	93.30$\pm$1.43	&	93.54$\pm$1.16	&	92.08$\pm$1.48	&	93.43$\pm$0.82	&	\textbf{93.56$\pm$1.24}	&	93.38$\pm$0.80	&	93.41$\pm$0.90	&	93.45$\pm$1.00	\\
Wisconsin   	&	94.37$\pm$3.58	&	94.67$\pm$3.25$\ddagger$	&	95.56$\pm$4.06	&	95.41$\pm$3.25	&	93.93$\pm$3.68	&	94.96$\pm$3.86$\ddagger$	&	95.26$\pm$3.50	&	95.41$\pm$3.79	&	95.56$\pm$3.70	&	\textbf{95.70$\pm$3.45}	\\
Credit	        &	100.00$\pm$0.00	&	100.00$\pm$0.00	&	100.00$\pm$0.00	&	100.00$\pm$0.00	&	100.00$\pm$0.00	&	100.00$\pm$0.00	&	100.00$\pm$0.00	&	100.00$\pm$0.00	&	100.00$\pm$0.00	&	100.00$\pm$0.00	\\
Landsat	        &	97.34$\pm$0.45	&	97.22$\pm$0.66	&	97.29$\pm$0.37	&	97.31$\pm$0.28	&	96.08$\pm$1.06$\ddagger$	&	97.26$\pm$0.26	&	97.34$\pm$0.35	&	97.29$\pm$0.56	&	97.15$\pm$0.43	&	\textbf{97.42$\pm$0.34}	\\
Wilt	        &	98.24$\pm$0.32	&	98.24$\pm$0.26	&	98.37$\pm$0.42	&	98.28$\pm$0.32	&	98.01$\pm$0.51	&	98.37$\pm$0.51	&	98.16$\pm$0.18	&	98.39$\pm$0.53	&	\textbf{98.41$\pm$0.33}	&	98.35$\pm$0.38	\\
Shuttle	        &	99.96$\pm$0.03	&	99.96$\pm$0.03	&	\textbf{99.97$\pm$0.02}	&	\textbf{99.97$\pm$0.02}	&	\textbf{99.97$\pm$0.02}	&	99.96$\pm$0.03	&	99.96$\pm$0.03	&	\textbf{99.97$\pm$0.02}	&	99.96$\pm$0.03	&	99.97$\pm$0.03	\\
Ecoli	        &	94.55$\pm$3.14	&	94.85$\pm$2.03	&	93.64$\pm$2.71	&	95.15$\pm$2.49	&	91.52$\pm$3.49$\ddagger$	&	94.85$\pm$2.75	&	93.03$\pm$2.30	&	94.24$\pm$2.71	&	95.15$\pm$2.49	&	\textbf{95.15$\pm$1.98}	\\
SensorReadings  &	99.25$\pm$0.40	&	99.19$\pm$0.50	&	99.41$\pm$0.47	&	\textbf{99.52$\pm$0.24}	&	99.28$\pm$0.25	&	99.45$\pm$0.23	&	99.50$\pm$0.19	&	99.36$\pm$0.43	&	99.43$\pm$0.33	&	99.38$\pm$0.41	\\
    \midrule
    $t$-Test (W/T/L) & 0/10/0 & 1/9/0 & 0/10/0 & 1/9/0 & 2/8/0 & 2/8/0 & 0/10/0 & 0/10/0 & 0/10/0 & --- \\
    Average Rank & 7.15 & 7.20 & 4.80 & 4.60 & 8.40 & 5.55 & 5.20 & 4.75 & 4.40 & 2.95 \\
    \bottomrule
\end{tabular}
\begin{tablenotes}
    \item[1] The reported results are the average test accuracy of each method and the corresponding standard deviation under 5-fold cross-validation on each data set. 
    The best results with higher accuracy and lower standard deviation are highlighted in boldface. 
    \item[2] By two-tailed paired $t$-test at 5\% significance level, 
    $\ddagger$ and $\dagger$ denote that the performance of \ddismi{} is superior to and inferior to that of the comparative method, respectively. 
    \item[3] The last two rows show the results of $t$-test and average rank, respectively. 
    The ``W/T/L'' in $t$-test indicates that \ddismi{} is superior to, not significantly different from, or inferior to the corresponding comparative methods. 
    The average rank is calculated according to the Friedman test~\cite{demvsar2006statistical}. 
\end{tablenotes}
\end{threeparttable}
}
\end{table*}

\begin{table*}[t]
\centering
\caption{
Comparison of the state-of-the-art methods with \greedy{} and \ddismi{} using Bagging to produce an ensemble with SVMs as individual classifiers.
}\label{tab:Bagging:SVM}
\scalebox{0.97}{
\setlength{\tabcolsep}{3.2pt}
\begin{tabular}{c cccccccc cc}
    \toprule
    Dataset & KL & KP & OO & RE & DREP & SEP & OEP & PEP & \greedy{} & \ddismi{} \\
    \midrule
Liver	        &	\textbf{58.84$\pm$1.65}	&	\textbf{58.84$\pm$1.65}	&	58.26$\pm$1.21	&	\textbf{58.84$\pm$1.65}	&	\textbf{58.84$\pm$1.65}	&	58.55$\pm$1.65	&	\textbf{58.84$\pm$1.65}	&	58.55$\pm$1.65	&	\textbf{58.84$\pm$1.65}	&	\textbf{58.84$\pm$1.65}	\\
Ringnorm	    &	98.43$\pm$0.37	&	\textbf{98.53$\pm$0.30}$\dagger$	&	98.51$\pm$0.36$\dagger$	&	98.47$\pm$0.29	&	98.43$\pm$0.29	&	98.47$\pm$0.36	&	98.39$\pm$0.25	&	98.44$\pm$0.31	&	98.43$\pm$0.31	&	98.43$\pm$0.31	\\
Waveform	    &	91.45$\pm$1.60	&	91.23$\pm$1.50	&	91.35$\pm$1.58	&	91.17$\pm$1.67	&	90.83$\pm$1.52$\ddagger$	&	91.27$\pm$1.53	&	91.13$\pm$1.61	&	91.37$\pm$1.63	&	\textbf{91.45$\pm$1.49}	&	91.23$\pm$1.63	\\
Credit	        &	78.02$\pm$0.07	&	78.01$\pm$0.11	&	78.03$\pm$0.08	&	78.02$\pm$0.08	&	77.96$\pm$0.12	&	\textbf{78.04$\pm$0.12}	&	78.02$\pm$0.11	&	78.02$\pm$0.13	&	78.00$\pm$0.09	&	78.01$\pm$0.11	\\
Landsat	        &	65.24$\pm$0.00	&	65.24$\pm$0.00	&	65.24$\pm$0.00	&	65.24$\pm$0.00	&	65.24$\pm$0.00	&	65.24$\pm$0.00	&	65.24$\pm$0.00	&	65.24$\pm$0.00	&	65.24$\pm$0.00	&	65.24$\pm$0.00	\\
Page	        &	91.26$\pm$0.41	&	91.26$\pm$0.45	&	91.28$\pm$0.44	&	91.24$\pm$0.46	&	91.28$\pm$0.52	&	91.22$\pm$0.32	&	91.24$\pm$0.44	&	\textbf{91.30$\pm$0.53}	&	91.26$\pm$0.45	&	91.22$\pm$0.44	\\
Wilt	        &	94.68$\pm$0.09	&	94.68$\pm$0.09	&	94.68$\pm$0.09	&	94.68$\pm$0.09	&	94.68$\pm$0.09	&	94.68$\pm$0.09	&	94.68$\pm$0.09	&	94.68$\pm$0.09	&	94.68$\pm$0.09	&	94.68$\pm$0.09	\\
SensorReadings  &	89.17$\pm$0.78	&	89.24$\pm$0.88	&	89.28$\pm$0.52	&	89.35$\pm$0.51	&	88.91$\pm$0.60	&	89.11$\pm$0.91$\ddagger$	&	89.26$\pm$0.50	&	\textbf{89.51$\pm$0.94}	&	89.42$\pm$0.83	&	89.37$\pm$1.04	\\
EEGEyeState	    &	55.13$\pm$0.00	&	55.13$\pm$0.00	&	55.13$\pm$0.00	&	55.13$\pm$0.00	&	55.13$\pm$0.00	&	55.13$\pm$0.00	&	55.13$\pm$0.01	&	55.13$\pm$0.00	&	55.13$\pm$0.00	&	55.13$\pm$0.00	\\
WaveformNoise	&	86.35$\pm$0.74	&	86.11$\pm$0.83	&	86.37$\pm$0.79	&	86.21$\pm$0.94	&	85.99$\pm$0.83	&	86.49$\pm$1.02	&	86.11$\pm$0.98	&	86.29$\pm$0.77	&	\textbf{86.55$\pm$0.68}	&	86.43$\pm$0.81	\\
    \midrule
    $t$-Test (W/T/L) & 0/10/0 & 0/9/1 & 0/9/1 & 0/10/0 & 1/9/0 & 1/9/0 & 0/10/0 & 0/10/0 & 0/10/0 & --- \\
    Average Rank & 5.20 & 5.60 & 4.60 & 5.50 & 7.05 & 5.50 & 6.60 & 4.55 & 4.65 & 5.75 \\
    \bottomrule
\end{tabular}
}
\end{table*}

\subsection{Comparison of \greedy{} and \ddismi{} to the State-of-the-art Ensemble Pruning Methods}
\label{cmp:centralized}

In this subsection, we compare the quality of various ensemble pruning methods (the original centralized version) including KL, KP, OO, RE, DREP, SEP, OEP, and PEP with our proposed centralized (\greedy) and distributed (\ddismi) methods. 
\begin{figure}[t]
\centering
\subfloat[]{\centering\label{trial1:cmprank}
	\includegraphics[scale=0.67]{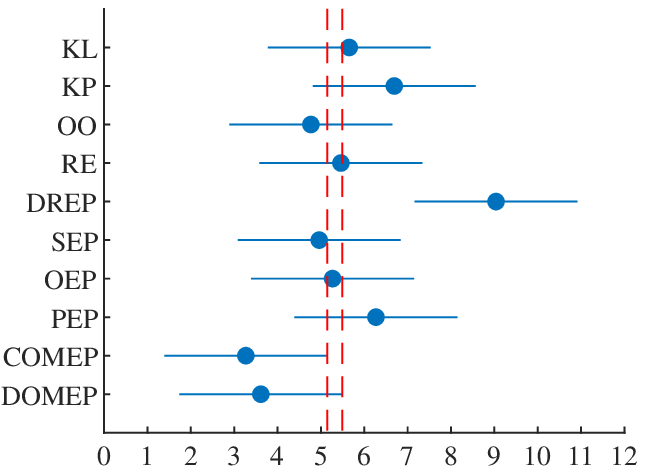}}
\hspace{0.2em}
\subfloat[]{\centering\label{trial1:demvsar}
	\includegraphics[scale=0.63]{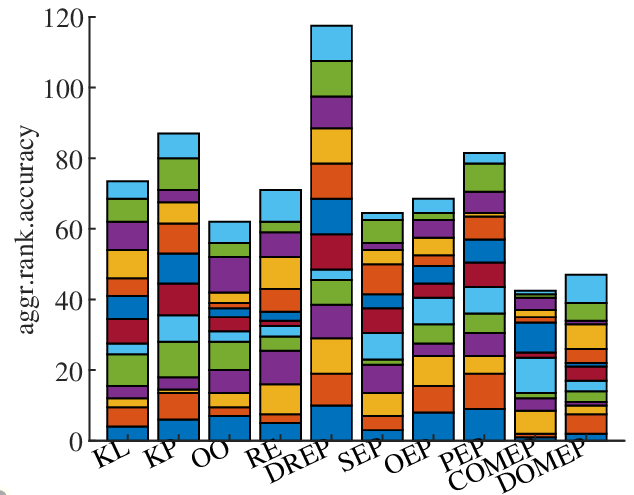}}
\caption{
Comparison of the state-of-the-art methods with \greedy{} and \ddismi{} on the test accuracy.
(a) Friedman test chart (non-overlapping means significant difference) \cite{demvsar2006statistical}. 
(b) The aggregated rank for each method (the smaller the better) \cite{qian2015pareto}. 
}\label{trial1}
\end{figure}

\begin{figure}[t]
\centering
    \subfloat[]{\label{fig:bagsvm,m2a}\centering
        \includegraphics[scale=0.65]{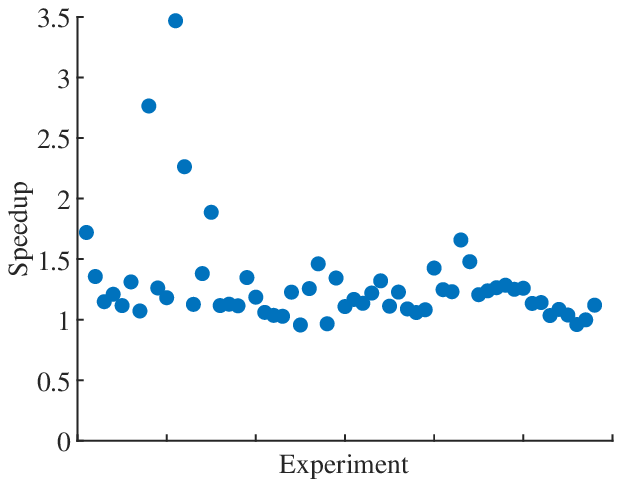}}
    \hspace{0.5em}
    \subfloat[]{\label{fig:bagsvm,m3a}\centering
        \includegraphics[scale=0.65]{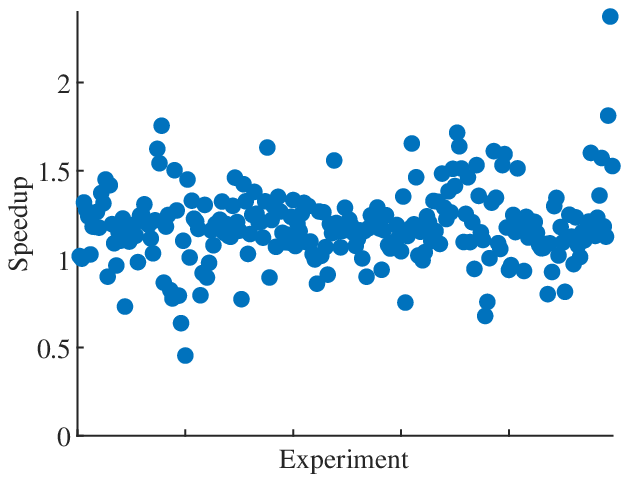}}
    \hspace{0.5em} 
    \subfloat[]{\label{fig:bagsvm,m2b}\centering
        \includegraphics[scale=0.65]{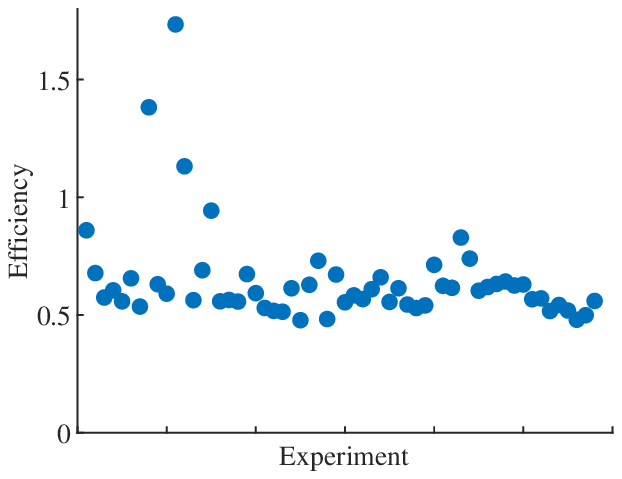}}
    \hspace{0.5em}
    \subfloat[]{\label{fig:bagsvm,m3b}\centering
        \includegraphics[scale=0.65]{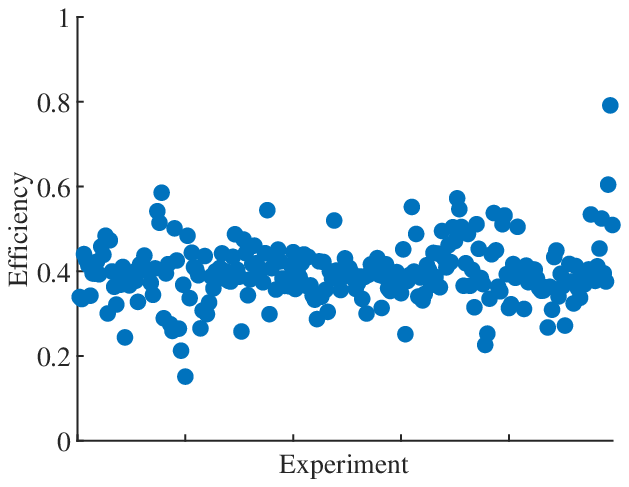}}
\caption{
Comparison of speedup and efficiency between \greedy{} and \ddismi{}. 
(a) Speedup with two machines. 
(b) Speedup with three machines. 
(c) Efficiency with two machines. 
(d) Efficiency with three machines. 
}\label{fig:Bag,SVM}
\end{figure}

Experimental results reported in Table~\ref{tab:Bagging:DT} contain the average test accuracy of each method and the corresponding standard deviation under 5-fold cross-validation on each data set. 
Each row in Table~\ref{tab:Bagging:DT} compares the classification accuracy using bagging with the same type of individual classifiers. 
The results with higher accuracy and lower standard deviation are indicated with bold fonts for each data set (row). 
Besides, we examine the significance of the difference in the accuracy performance between two ensemble pruning methods by two-tailed paired $t$-test at 5\% significance level to tell whether these two methods have significantly different results. 
Two methods end in a tie if there is no significant statistical difference; otherwise, the one with higher values of accuracy will win. 
In the last two rows of Table~\ref{tab:Bagging:DT}, the average rank of each method is calculated based on the Friedman test~\cite{demvsar2006statistical}, and the performance of each method is compared with \ddismi{} in terms of the number of data sets that \ddismi{} has won, tied, or lost, respectively. 
It can be inferred that \ddismi{} does not underperform centralized methods in many data sets even though it only utilizes local information unlike others, which confirms the reasonableness of \ddismi{} (and \greedy{}) utilizing accuracy and diversity simultaneously. 
Despite slightly lower values of accuracy in some cases, \ddismi{} remains acceptable results. 
Similar results are reported in Table~\ref{tab:Bagging:SVM} and Table~\ref{tab:Bagging:KNNd} using different individual classifiers. 
Moreover, Figure~\ref{trial1} reports the comparison of the state-of-the-art methods with \greedy{} and \ddismi{} on the test accuracy using statistical test methods~\cite{demvsar2006statistical,qian2015pareto}. 
Figure~\ref{trial1}\subref{trial1:cmprank} shows that \greedy{} and \ddismi{} have significant superiority over other compared centralized methods, utilizing the Friedman test chart~\cite{demvsar2006statistical}; 
Figure~\ref{trial1}\subref{trial1:demvsar} represents the aggregated rank for each method, depicting the same results.

\begin{table*}
\centering
\caption{
Comparison of the state-of-the-art methods with \greedy{} and \ddismi{} using Bagging to produce an ensemble with KNNs as individual classifiers.
}\label{tab:Bagging:KNNd}
\scalebox{0.97}{
\setlength{\tabcolsep}{3.2pt}
\begin{tabular}{c cccccccc cc}
    \toprule
    Dataset & KL & KP & OO & RE & DREP & SEP & OEP & PEP & \greedy{} & \ddismi{} \\
    \midrule
Ames	        &	58.83$\pm$5.48	&	58.54$\pm$7.20	&	\textbf{60.29$\pm$7.45}	&	57.81$\pm$7.14	&	57.52$\pm$7.29	&	59.71$\pm$6.18	&	59.56$\pm$6.96	&	58.69$\pm$5.90	&	60.15$\pm$6.90	&	59.12$\pm$5.95	\\
Card	        &	67.59$\pm$2.10	&	66.86$\pm$1.68	&	66.72$\pm$1.68	&	67.01$\pm$2.27	&	64.38$\pm$2.27	&	67.88$\pm$2.87	&	65.99$\pm$1.22	&	65.55$\pm$2.39$\ddagger$	&	\textbf{68.32$\pm$3.33}	&	68.03$\pm$3.44	\\
Sonar	        &	80.98$\pm$7.98	&	80.00$\pm$8.52	&	81.95$\pm$7.03	&	81.95$\pm$5.62	&	79.02$\pm$5.62	&	81.46$\pm$5.87	&	80.00$\pm$5.29	&	78.05$\pm$5.72	&	\textbf{82.44$\pm$7.19}	&	80.98$\pm$8.69	\\
Page	        &	95.87$\pm$0.35	&	95.87$\pm$0.49	&	95.81$\pm$0.45	&	95.76$\pm$0.35	&	95.76$\pm$0.39	&	95.80$\pm$0.57	&	95.87$\pm$0.55	&	95.81$\pm$0.46	&	95.87$\pm$0.64	&	\textbf{95.89$\pm$0.64}	\\
Wilt	        &	97.93$\pm$0.28	&	97.89$\pm$0.32	&	\textbf{98.01$\pm$0.24}	&	97.99$\pm$0.20	&	97.77$\pm$0.36	&	97.93$\pm$0.35	&	97.89$\pm$0.40	&	98.01$\pm$0.32	&	97.97$\pm$0.19	&	98.01$\pm$0.26	\\
Landsat	        &	90.28$\pm$0.88	&	90.22$\pm$1.18	&	90.30$\pm$0.95	&	90.40$\pm$0.83	&	90.31$\pm$0.94	&	\textbf{90.47$\pm$0.78}	&	90.37$\pm$0.85	&	90.37$\pm$0.76	&	90.47$\pm$1.21	&	90.44$\pm$0.87	\\
Shuttle	        &	99.82$\pm$0.03	&	99.81$\pm$0.03	&	99.82$\pm$0.04	&	\textbf{99.82$\pm$0.02}	&	99.82$\pm$0.04	&	99.81$\pm$0.04	&	99.81$\pm$0.03	&	99.81$\pm$0.03	&	99.80$\pm$0.03$\ddagger$	&	99.82$\pm$0.04	\\
Ecoli	        &	95.15$\pm$2.91	&	94.55$\pm$3.14	&	\textbf{95.76$\pm$2.91}	&	95.45$\pm$3.03	&	93.33$\pm$3.14	&	94.85$\pm$2.95	&	94.24$\pm$3.11	&	94.85$\pm$3.65	&	95.45$\pm$3.03	&	\textbf{95.76$\pm$2.91}	\\
WaveformNoise	&	84.34$\pm$1.05	&	84.20$\pm$1.46	&	84.48$\pm$2.09	&	84.36$\pm$2.24	&	83.54$\pm$0.74	&	84.36$\pm$1.33	&	84.48$\pm$2.24	&	84.16$\pm$1.58	&	84.48$\pm$1.81	&	\textbf{84.62$\pm$1.77}	\\
EEGEyeState	    &	95.99$\pm$0.55	&	96.06$\pm$0.52	&	96.16$\pm$0.50	&	95.98$\pm$0.59	&	95.03$\pm$0.37$\ddagger$	&	96.11$\pm$0.50	&	96.08$\pm$0.66	&	\textbf{96.28$\pm$0.55}	&	96.19$\pm$0.62	&	96.01$\pm$0.46	\\
SensorReadings  &	96.29$\pm$0.47	&	96.11$\pm$0.39	&	96.33$\pm$0.62	&	96.37$\pm$0.53	&	96.00$\pm$0.51	&	96.29$\pm$0.64	&	96.39$\pm$0.42	&	96.28$\pm$0.60	&	\textbf{96.42$\pm$0.64}	&	96.31$\pm$0.31	\\
    \midrule
    $t$-Test (W/T/L) & 0/11/0 & 0/11/0 & 0/11/0 & 0/11/0 & 1/10/0 & 0/11/0 & 0/11/0 & 1/10/0 & 1/10/0 & --- \\
    Average Rank & 5.82 & 7.45 & 3.77 & 5.27 & 8.95 & 5.09 & 5.77 & 6.55 & 3.05 & 3.27 \\
    \bottomrule
\end{tabular}
}
\end{table*}

\begin{figure}[t]
\centering
\begin{minipage}{\linewidth}
    \centering
    \subfloat[]{\label{fig:BagDT:bisensor24:a}\centering
        \includegraphics[scale=0.68]{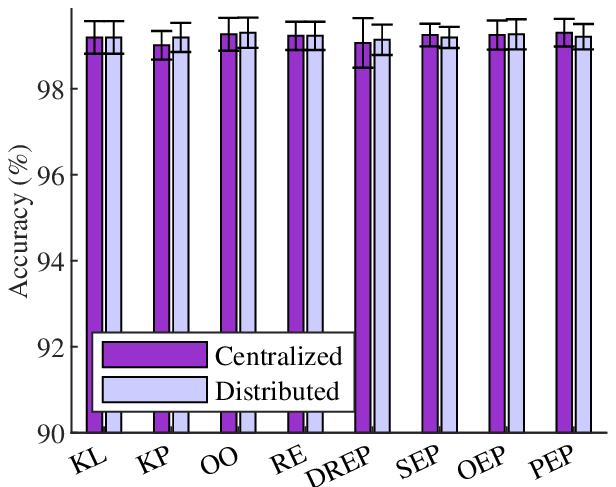}}
    \hspace{0.5em}
    \subfloat[]{\label{fig:BagDT:bisensor24:b}\centering
        \includegraphics[scale=0.68]{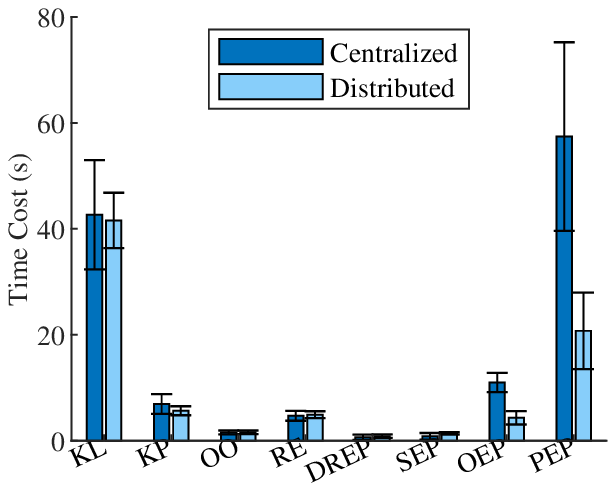}}
\caption{
Comparison between the state-of-the-art ensemble pruning methods and their distributed versions on the SensorReadings dataset using Bagging with DTs as individual classifiers for binary classification. 
(a) Accuracy. (b) Time Cost. 
}\label{fig:Bag,DT:bi,sensor24}
\end{minipage}
\begin{minipage}{\linewidth}
\vspace{1.25em}
\centering
    \subfloat[]{\label{fig:BagKNNu:bisensor2:a}\centering
        \includegraphics[scale=0.68]{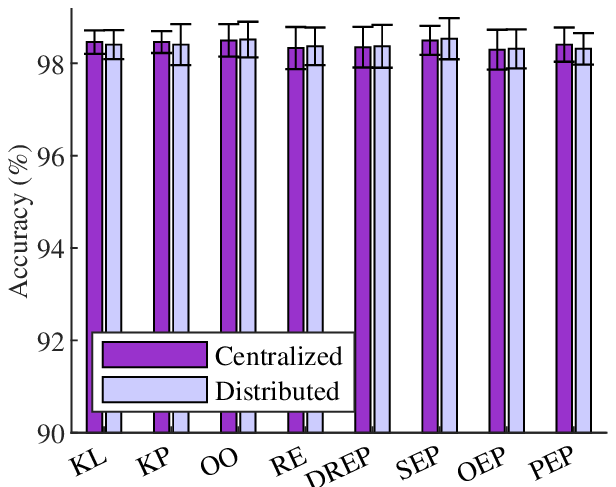}}
    \hspace{0.5em}
    \subfloat[]{\label{fig:BagKNNu:bisensor2:b}\centering
        \includegraphics[scale=0.68]{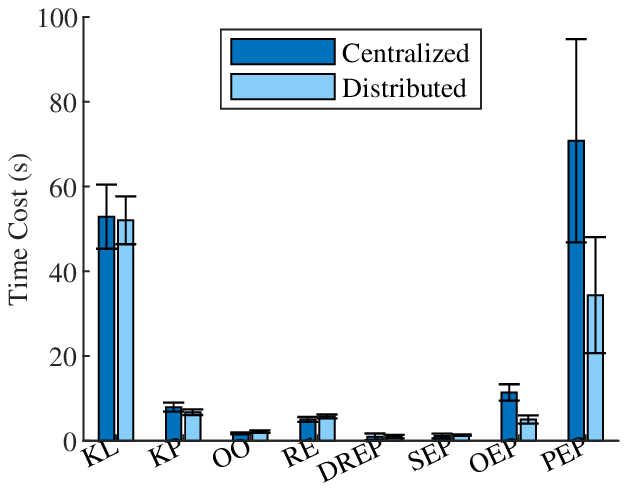}}
\caption{
Comparison between the state-of-the-art ensemble pruning methods and their distributed versions on the SensorReadings dataset using Bagging with KNNs as individual classifiers for binary classification. 
(a) Accuracy. (b) Time Cost. 
}\label{fig:Bag,KNNu:bi,sensor2}
\end{minipage}
\end{figure}
%
\begin{figure}[t]
\centering
\begin{minipage}{\linewidth}
\centering
    \subfloat[]{\label{fig:BagDT:musensor24:a}\centering
        \includegraphics[scale=0.68]{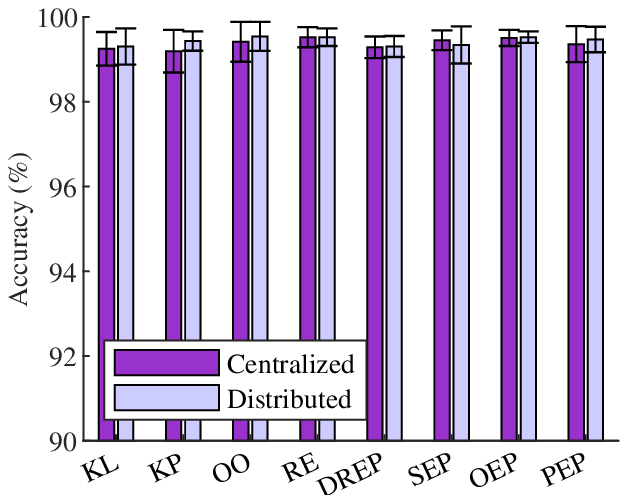}}
    \hspace{0.5em}
    \subfloat[]{\label{fig:BagDT:musensor24:b}\centering
        \includegraphics[scale=0.68]{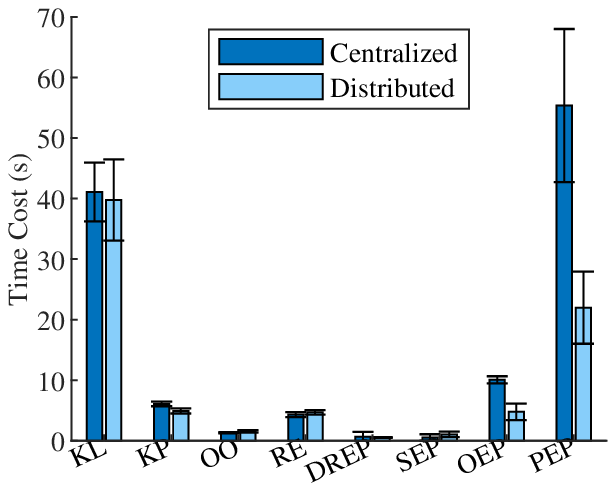}}
\caption{
Comparison between the state-of-the-art ensemble pruning methods and their distributed versions on the SensorReadings dataset using Bagging with DTs as individual classifiers for multi-class classification. 
(a) Accuracy. (b) Time Cost. 
}\label{fig:Bag,DT:mu,sensor24}
\end{minipage}
\begin{minipage}{\linewidth}
\vspace{1.25em}
\centering
    \subfloat[]{\label{fig:BagKNNu:musensor4:a}\centering
        \includegraphics[scale=0.68]{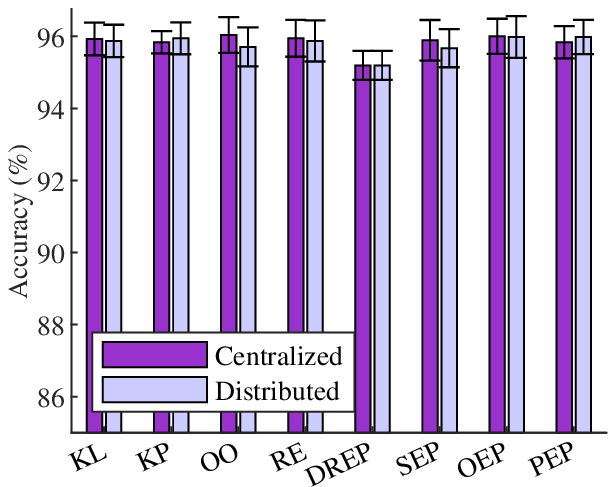}}
    \hspace{0.5em}
    \subfloat[]{\label{fig:BagKNNu:musensor4:b}\centering
        \includegraphics[scale=0.68]{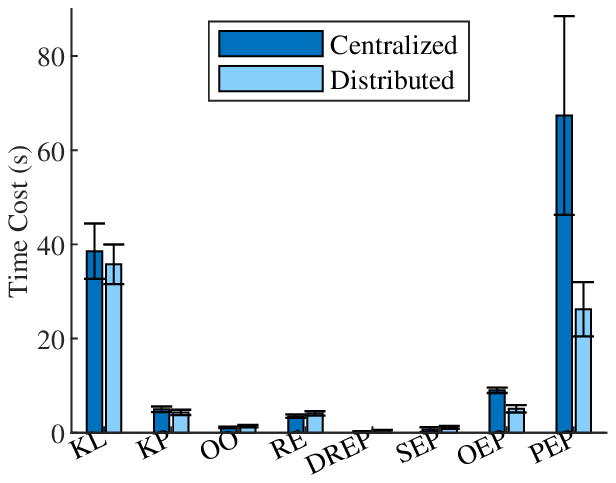}}
\caption{
Comparison between the state-of-the-art ensemble pruning methods and their distributed versions on the SensorReadings dataset using Bagging with KNNs as individual classifiers for multi-class classification. 
(a) Accuracy. (b) Time Cost. 
}\label{fig:Bag,KNNu:mu,sensor4}
\end{minipage}
\end{figure}

\subsection{\ddismi{} vs. \greedy{} over the Time Cost} 
\label{cmp:mynew}

In this subsection, we give the complexity analysis of the proposed methods and present the corresponding experimental results. 
According to the Algorithm~\ref{greedy}, the computational complexity of \greedy{} is analyzed as follows: 
\begin{itemize}
    \item Firstly, the complexity of an essential term of $\dist(h_i,h_j)$ is $\mathcal{O}(d^2n_c^2)$, where $d$ and $n_c$ are the number of instances and labels, respectively. 
    \item Secondly, the chosen individual classifier is obtained with the complexity of $\mathcal{O}(-\frac{1}{3}k^3+\frac{1}{2}nk^2)$ times that of the $\dist(h_i,h_j)$ term since $\sum_{i=2}^k(i-1)(n-i+2)= -\frac{1}{3}k^3+\frac{n+2}{2}k^2-\frac{3n+4}{6}k$, where $k$ is the size of the pruned sub-ensemble. 
\end{itemize}
Therefore, the overall computational complexity of \greedy{} is $\mathcal{O}\big( (-\frac{1}{3}k^3+\frac{1}{2}nk^2)d^2n_c^2 \big)$. 
Besides, since $\mathcal{H}$ is partitioned into $m$ groups in \ddismi{}, the overall computational complexity of \ddismi{} is $\mathcal{O}\big( (-\frac{1}{3}k^3+\frac{n}{2m}k^2)d^2n_c^2 \big)$.

In this experiment, we employ different numbers of machines in \ddismi{} in order to test its speedup in comparison with \greedy{}. 
Under the ideal conditions, Zadeh \etal{}~\cite{zadeh2017scalable} pointed that the speedup between the distributed and centralized version was almost linear in terms of the number of used machines since there was no overhead of information-sharing between those machines. 
To verify whether \ddismi{} could achieve competitive performance faster than \greedy{} or not, we introduce two performance indicators from parallel processing, \ie{} speedup and efficiency\footnote{%
Efficiency is a measure of how effectively parallel computing could be used to solve a particular problem. 
A parallel algorithm is considered cost efficient if its asymptotic running time multiplied by the number of processing units involved in the computation is comparable to the running time of the best sequential algorithm \cite{wiki_speedup}. 
}, where speedup is defined as a quotient of the execution time of \greedy{} and that of \ddismi{}, to report how much speedup of \ddismi{} over \greedy{}. 
Constrained by the capability of machines on which we test, several experiments for ensemble pruning conducted on two or three machines are used as a typical example to present the performance of \ddismi{}. 
The results using various settings of this experiment are summarized in Figure~\ref{fig:Bag,SVM}, which indicates that \ddismi{} runs faster than \greedy{} even reaching super-linear speedup\footnote{%
Sometimes a speedup of more than $m$ when using $m$ processors is observed in parallel computing, which is called super-linear speedup \cite{wiki_speedup}. 
} on a tiny minority of experiments.

\subsection{Comparison Between the State-of-the-art Ensemble Pruning Methods and Their Corresponding Distributed Versions Generated with \distributed{}}
\label{cmp:distributed}

In this subsection, to verify the effectiveness of \distributed{}, we compare the quality of various centralized ensemble pruning methods and their respective distributed versions that are generated using \distributed{} in terms of accuracy and time cost. 
To test the quality of the selected sub-ensembles of each method, we control them under the same conditions (including the employed ensemble methods or types of individual classifiers) in each experiment. 
Figure~\ref{fig:Bag,DT:bi,sensor24} shows the comparison results when individual classifiers are designated as DTs and assembled by bagging for binary classification. 
It can be inferred that, for each pruning method (each group on the horizontal axis), the accuracy of the distributed version is superior or equal to that of its corresponding centralized version. 
In consideration of the less time cost it takes, we believe that the distributed version of each method outperforms its original centralized version. 
Besides, we can tell that the effectiveness of \distributed{} is exceptionally evident on PEP, \ie{} a complicated method utilizing an evolutionary Pareto optimization combined with a local search subroutine. 
Moreover, Figure~\ref{fig:Bag,KNNu:bi,sensor2} reports similar results of that with KNNs as individual classifiers for binary classification; Figures~\ref{fig:Bag,DT:mu,sensor24}--\ref{fig:Bag,KNNu:mu,sensor4} report similar results of that with DTs and KNNs as individual classifiers for multi-class classification, respectively.

\begin{figure}[t]
\centering
\begin{minipage}{\linewidth}
\centering
    \subfloat[]{\label{fig:eg4:Bag_DT:Ames1}\centering
        \includegraphics[scale=0.69]{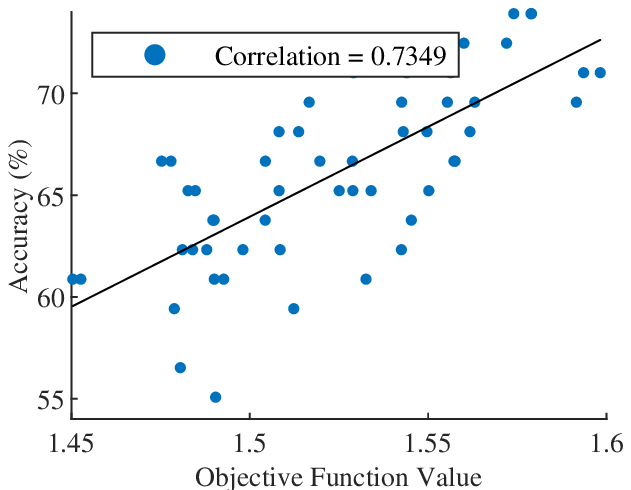}}
    \vspace{0.5em}
    \subfloat[]{\label{fig:eg4:Bag_DT:Ames2}\centering
        \includegraphics[scale=0.68]{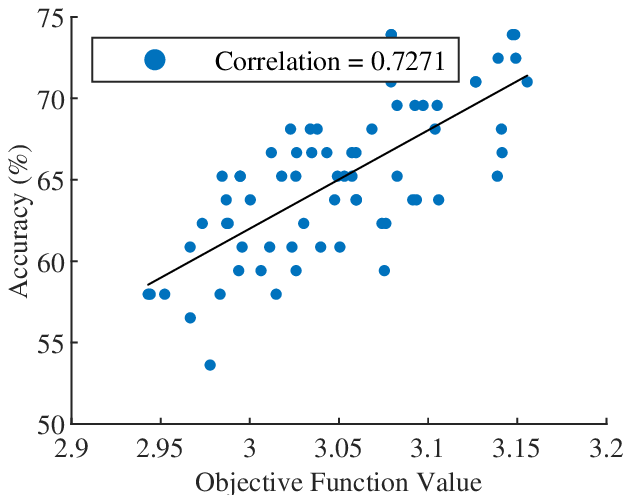}}
\caption{
Relation of binary classification accuracy and objective function value for 3-combinations and 4-combinations in the Ames dataset using Bagging with DTs as individual classifiers for binary classification. 
(a) 3-combinations. (b) 4-combinations. 
}\label{fig:eg4:Bag_DT:Ames}
\end{minipage}
\begin{minipage}{\linewidth}
\vspace{1.25em}
\centering
    \subfloat[]{\label{fig:eg4:Bag_DT:wave2}\centering
        \includegraphics[scale=0.68]{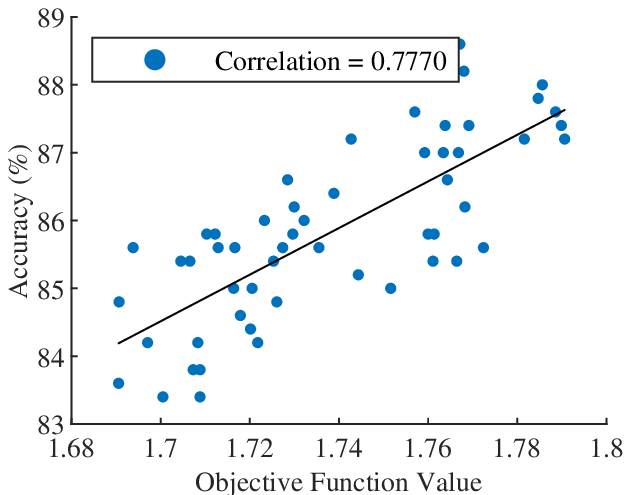}}
    \vspace{0.5em}
    \subfloat[]{\label{fig:eg4:Bag_DT:wave3}\centering
        \includegraphics[scale=0.68]{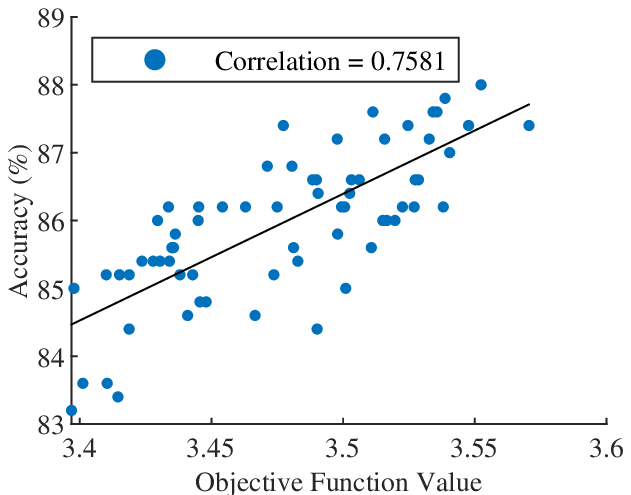}}
\caption{
Relation of binary classification accuracy and objective function value for 3-combinations and 4-combinations in the Waveform dataset using Bagging with DTs as individual classifiers for binary classification. 
(a) 3-combinations. (b) 4-combinations. 
}\label{fig:eg4:Bag_DT:wave}
\end{minipage}
\begin{minipage}{\linewidth}
\vspace{1.25em}
\centering
    \subfloat[]{\label{fig:eg4:Bag_DT:wavemu1}\centering
        \includegraphics[scale=0.68]{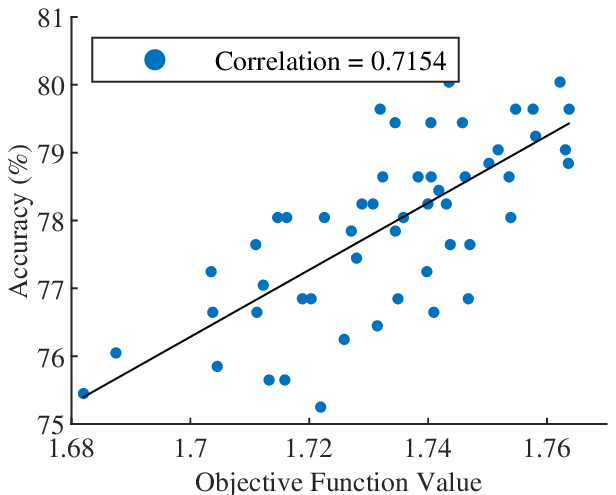}}
    \vspace{0.5em}
    \subfloat[]{\label{fig:eg4:Bag_DT:wavemu2}\centering
        \includegraphics[scale=0.68]{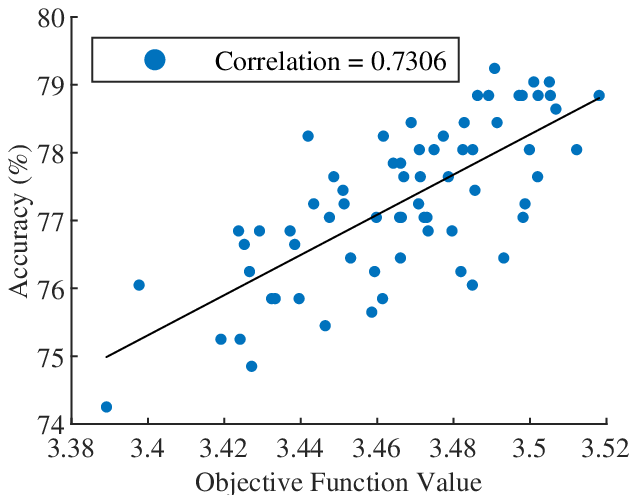}}
\caption{
Relation of multi-class classification accuracy and objective function value for 3-combinations and 4-combinations in the Waveform dataset using Bagging with DTs as individual classifiers for multi-classification. 
(a) 3-combinations. (b) 4-combinations. 
}\label{fig:eg4:Bag_DT:wavemu}
\end{minipage}
\end{figure}

\subsection{Validating the Objective Function}

Regarding a new objective function, its relation with classification accuracy is one of the fundamental questions. 
We select two small-sized ensembles (small in the number of individual classifiers) and evaluate all possible combinations of these individual classifiers in order to test this issue. 
In this experiment, we compare the classification accuracy for all the 3-combinations and 4-combinations of individual classifiers in the original ensemble against their corresponding objective value with the $\lambda$ parameter equal to 0.5, which means that two criteria in Eq.~(\ref{eq:4}) are equally important. 
Each small blue dot in Figure~\ref{fig:eg4:Bag_DT:Ames} represents the classification accuracy on a 3-combination or 4-combination of the individual classifiers with the size 8 of an ensemble in the Ames dataset for binary classification, and the line is the regression line. 
Similar results are shown in Figures~\ref{fig:eg4:Bag_DT:wave}--\ref{fig:eg4:Bag_DT:wavemu} in the Waveform dataset for binary classification and multi-class classification, respectively. 
We observe that the objective value and the classification accuracy are highly correlated from Figures~\ref{fig:eg4:Bag_DT:Ames}--\ref{fig:eg4:Bag_DT:wavemu}, which means that maximizing this objective function leads to our target (\ie{} the highly accurate sub-ensembles).

\begin{figure}[t]
\centering
\begin{minipage}{\linewidth}
\centering
    \subfloat[]{\label{fig:eg5:BagSVM:biring:a}\centering
        \includegraphics[scale=0.66]{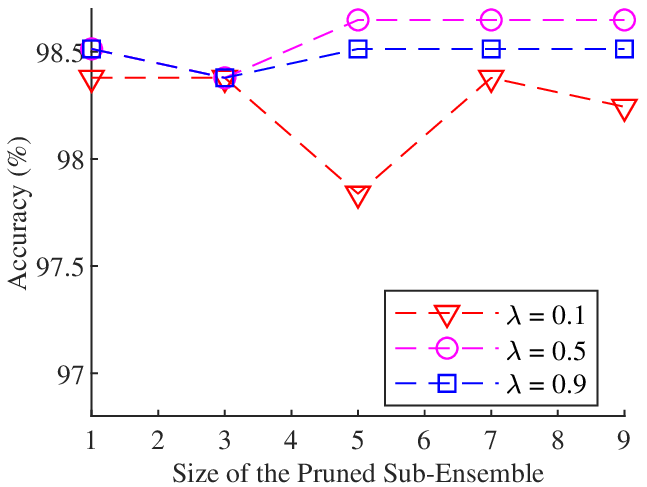}}
    \hspace{0.5em}
    \subfloat[]{\label{fig:eg5:BagSVM:biring:b}\centering
        \includegraphics[scale=0.66]{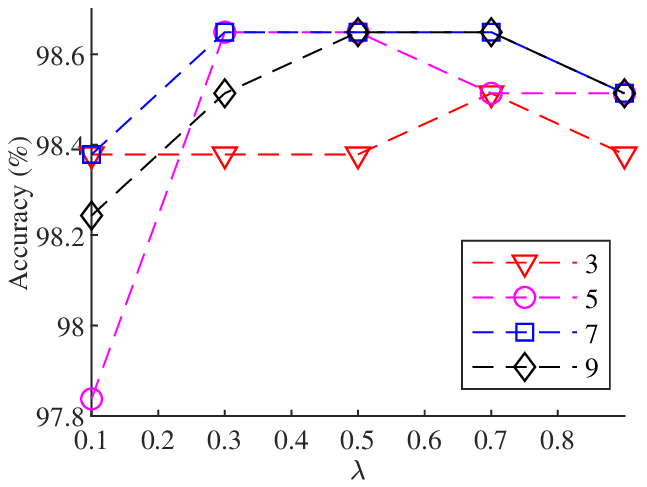}}
\caption{
Effect of $\lambda$ value on the classification accuracy in the Ringnorm dataset using Bagging with SVMs as individual classifiers for binary classification.
(a) Accuracy of each criterion individually. 
(b) Slight differences of $\lambda$ value while selecting the different size of the pruned sub-ensemble (3, 5, 7, 9).
}\label{fig:eg5:Bag,SVM:bi,ring}
\end{minipage}
\begin{minipage}{\linewidth}
\vspace{1.25em}
\centering
    \subfloat[]{\label{fig:eg5:BagKNNu:biheart:a}\centering
        \includegraphics[scale=0.66]{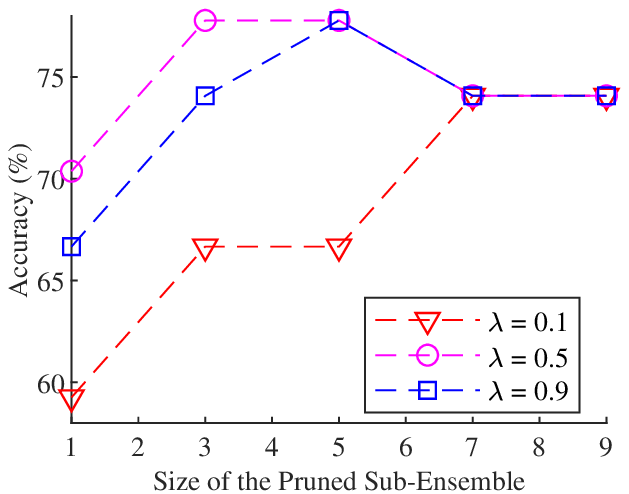}}
    \hspace{0.5em}
    \subfloat[]{\label{fig:eg5:BagKNNu:biheart:b}\centering
        \includegraphics[scale=0.66]{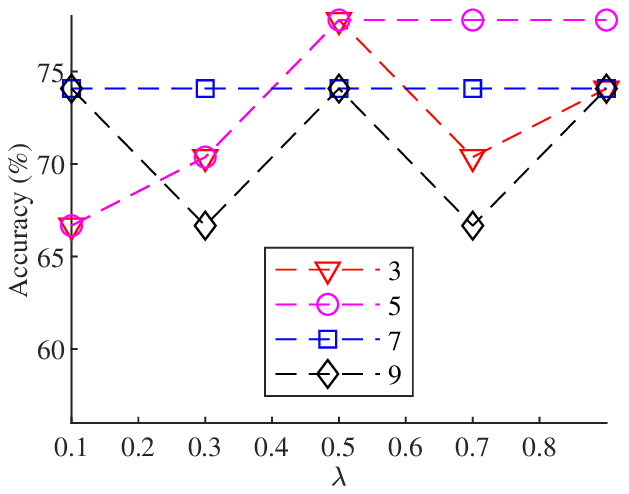}}
\caption{
Effect of $\lambda$ value on the classification accuracy in the Heart dataset using Bagging with KNNs as individual classifiers for binary classification.
(a) Accuracy of each criterion individually. 
(b) Slight differences of $\lambda$ value while selecting the different size of the pruned sub-ensemble (3, 5, 7, 9).
}\label{fig:eg5:Bag,KNNu:bi,heart}
\end{minipage}
\begin{minipage}{\linewidth}
\vspace{1.25em}
\centering
    \subfloat[]{\label{fig:eg5:BagDT:muwavenoise:a}\centering
        \includegraphics[scale=0.66]{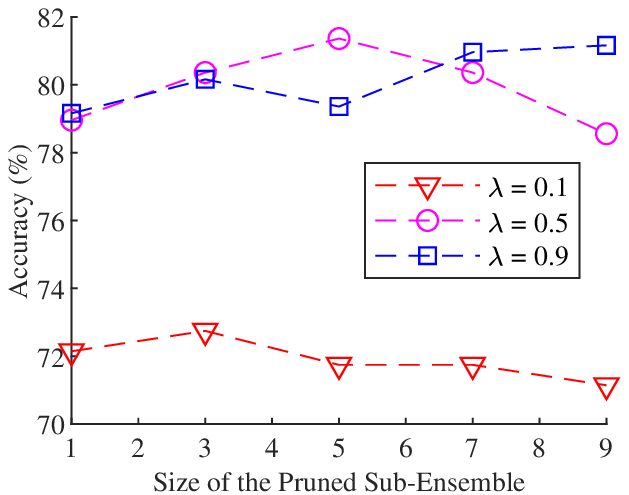}}
    \hspace{0.5em}
    \subfloat[]{\label{fig:eg5:BagDT:muwavenoise:b}\centering
        \includegraphics[scale=0.66]{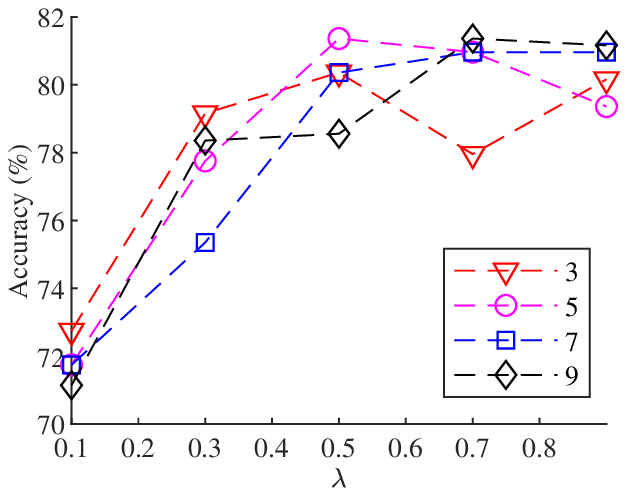}}
\caption{
Effect of $\lambda$ value on the classification accuracy in the Waveform dataset using Bagging with DTs as individual classifiers for multi-class classification.
(a) Accuracy of each criterion individually. 
(b) Slight differences of $\lambda$ value while selecting the different size of the pruned sub-ensemble (3, 5, 7, 9).
}\label{fig:eg5:Bag,DT:mu,wavenoise}
\end{minipage}
\end{figure}

\subsection{Effect of $\lambda$ Value}

Crucial as other issues, the relation of two criteria needs to be investigated in the defined objective function. 
To reveal how the classification results are affected with the regularization factor $\lambda \,,$ different $\lambda$ values (from 0.1 to 0.9 with 0.2 steps) are tested in the experiments of this part. 
Figure~\ref{fig:eg5:Bag,SVM:bi,ring} exemplifies the effect of $\lambda$ on the Ringnorm dataset.
Figure~\ref{fig:eg5:Bag,SVM:bi,ring}\subref{fig:eg5:BagSVM:biring:a} illustrates that the linear combination concurrently considering them both performs better than focusing more on $\mi$ term ($\lambda=0.1$) or $\vi$ term ($\lambda=0.9$) in Eq.~(\ref{divs2}) individually, although finding the optimal value of the $\lambda$ is another challenge. 
Figure~\ref{fig:eg5:Bag,SVM:bi,ring}\subref{fig:eg5:BagSVM:biring:b} presents that a global maximum around the optimal $\lambda$ exists indeed regardless of the size of the pruned sub-ensemble, which suggests that it might be related to the intrinsic properties of the data set. 
Similar results are reported in Figures~\ref{fig:eg5:Bag,KNNu:bi,heart}--\ref{fig:eg5:Bag,DT:mu,wavenoise} for binary classification and multi-class classification, respectively. 
Although proper results for all data sets have been brought with $\lambda$ being set to 0.5 (in Tables~\ref{tab:Bagging:DT}--\ref{tab:Bagging:KNNd}) for convenience, \ddismi{} would achieve a better performance in practice when the $\lambda$ is adjusted for each data set separately.

\section{Conclusion}
\label{result}

In this work, we formalize the ensemble pruning problem as an objection maximization problem based on information entropy to consider diversity and accuracy simultaneously. 
Then, we propose an ensemble pruning method according to this objection maximization problem (including two versions, \greedy{} and \ddismi{}) for ensemble pruning.
We also present that our methods (\greedy{} and \ddismi{}) are consistently competitive with various existing methods for ensemble pruning, which could handle large-scale ensembles fast yet efficiently through handling the accuracy and diversity of the ensembles properly. 
At last, we propose a general distributed framework (\distributed) for ensemble pruning, which could be widely applied to various existing methods for ensemble pruning, to achieve less time consuming without much accuracy degradation. 
The remarkable effectiveness of \distributed{} is positively valuable for enormous data in the real world.
For future work, it seems like a promising direction to explore the deeper theoretical basis and to try other objective functions to achieve better performance.


%

\ifCLASSOPTIONcaptionsoff
  \newpage
\fi



\bibliographystyle{IEEEtran}
\bibliography{title_abbr,refs,attached}

\end{document}